\PassOptionsToPackage{table}{xcolor} 
\documentclass[runningheads]{llncs}

\usepackage[T1]{fontenc}
\usepackage{graphicx}
\usepackage{listings}
\usepackage{todonotes}
\usepackage{xurl}
\usepackage{float}
\usepackage{hyperref}
\usepackage{tabularx}
\usepackage{placeins}

\begin{document}

\title{Safeguards for Speech2Speech LLM-Assistants:\\A Case Study in Automotive Applications}
\titlerunning{Safeguards for Speech2Speech LLM-Assistants}

\author{Gregor~Endler\inst{1}\orcidID{0009-0000-7405-4990} \and
        Sebastian~Kraus\inst{1} \and
        Lukas~Stappen\inst{2}
        }

\authorrunning{G. Endler et al.}
\institute{codemanufaktur GmbH, Germany, 
\email{kontakt@codemanufaktur.com}
\and
BMW Group, Germany\\
\url{http://www.codemanufaktur.com}, \url{http://www.bmwgroup.com}}

\maketitle

\begin{abstract}
Recent advances have introduced speech-to-speech (S2S) conversational assistants capable of producing natural-sounding interactions, including non-verbal cues like tonality and mood.
In the automotive domain, this enables intuitive and humanlike in-car dialogue experiences.
However, integrating these end-to-end assistants limits architectural options for programmable domain-specific safeguards.
This paper discusses two implementation approaches for S2S guardrails: transcript-based and tool-based.
Through an empirical evaluation, we demonstrate that both strategies are insufficient for industrial deployment in most cases due to prohibitive latency (delaying each answer by 0 to 1.4 seconds even for computationally cheap checks) and technical impediments (like potentially non-deterministic tool call behavior).
Finally, we outline open challenges for S2S guardrails in the automotive context.

\keywords{LLM \and Speech-to-Speech Assistants \and LLM Guardrails \and Responsible AI}
\end{abstract}

\section{Introduction}
Voice interaction is a natural fit for automotive applications since it reduces manual and visual workload and allows eyes on road and hands on wheel.
In contrast to traditional speech recognition systems, the use of Large Language Models (LLMs) in speech assistants enables free conversation flow instead of fixed keyword detection and a more natural understanding of user commands.
However, this conversational freedom also enables novel attacks on the  overall system.
This is exacerbated with Speech2Speech (S2S) systems:
While they offer new capabilities like producing and reacting to speech tonality, this comes at the cost of less control over the system's data flow:
S2S systems allow less control over intermediate results than traditional text-based systems, i.e. more steps are handed off to the model itself or to a vendor's black box.
Present safeguards do protect certain parts of such a system, but new attack surfaces also require new defenses.
It remains an open question whether vendor blackbox guardrail approaches can fully capture application specific requirements, organizational constraints, and contextual threat models.
Therefore, custom application-level guardrails are necessary to enforce concrete behavioral constraints beyond model intrinsic safeguards.

We explore two approaches for custom guardrails within implementation frameworks given by three major LLM vendors (Amazon, Google, and OpenAI): transcript-based and tool-based guardrails (see section \ref{sec:implementation_options}).
To verify their fundamental applicability, we performed latency tests comparing a no-protections baseline against both guardrail variants.
These experiments reveal that for two of the investigated APIs, guardrails introduce significant delays incompatible with fluid conversation, and that essential features for reliable guardrails integration are lacking.
We discuss directions for future work and current shortcomings that need to be addressed for in-production use of custom made guardrails in S2S assistants.

\section{Related Work\label{section:related_work}}
Many voice assistants rely on a chained architecture consisting of Speech Recognition, a text-based Large Language Model, and Text-to-Speech synthesis \cite{Cui2025}.
Recent Speech-to-Speech (S2S) architectures supersede this by processing and generating speech and audio directly without intermediate text \cite{Cui2025,peng2025survey,arora2025landscape}.
This has the advantage of more natural conversation and may include non-verbal speech features like emotion and tonality.
Multiple open- \cite{Defossez2024moshi,xu2025qwen2,long2025vita,ding2025kimi} and closed source solutions \cite{gpt-realtime,nova-sonic,gemini-flash}
exist.

Many approaches for LLM guardrails have been suggested \cite{Dong2024}.
Commercial providers offer configurable guardrails \cite{amazon-guardrails,openai-guardrails,google-guardrails}.
However, these are predominantly designed for text-based interactions and offer limited configurability for domain-specific requirements.
Some frameworks \cite{nvidia-guardrails,guardrailsai-guardrails} do
provide programmable guardrails, but their integration with streaming S2S APIs introduces latency that may be unacceptable for real-time conversation.

Deploying speech assistants in vehicles introduces domain-specific requirements spanning technical resilience, safety, privacy, and fairness \cite{Papagiannidis2025}.
Resilient systems must defend against attacks that could trigger dangerous vehicle maneuvers. 
Privacy concerns arise when attackers extract sensitive information such as calendar entries or navigation history \cite{Yan2022}.
Circumventing safeguards to generate hate speech or discriminatory content violates fairness principles \cite{Strohmier2024}.
Automotive standards like TARA (ISO/SAE 21434 \cite{standard2021iso}) were developed for traditional cyber-physical systems with deterministic behavior, not for the probabilistic, adaptive, and linguistically unpredictable characteristics of generative AI.
In addition, the automotive context imposes constraints including latency requirements for natural conversation (below approximately 0.4 seconds added delay \cite{brady1971,krauss1967,Schoenenberg2014}) and integration with safety-critical vehicle systems.
These constraints make direct application of generic guardrail frameworks challenging.

\section{S2S Guardrail Implementation Options\label{sec:implementation_options}}
\begin{figure}
\vspace{-0.6cm}
\centering
\begin{minipage}{.5\textwidth}
  \centering
  \includegraphics[width=0.74\linewidth]{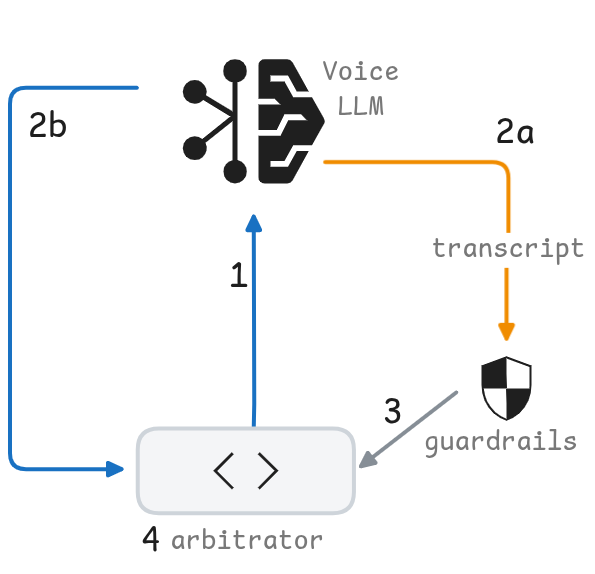}
  \vspace{-0.4cm}
  \caption{Transcript-Based Guardrails}
  \label{figure:transcript_guardrails}
\end{minipage}%
\begin{minipage}{.5\textwidth}
  \centering
  \includegraphics[width=0.95\linewidth]{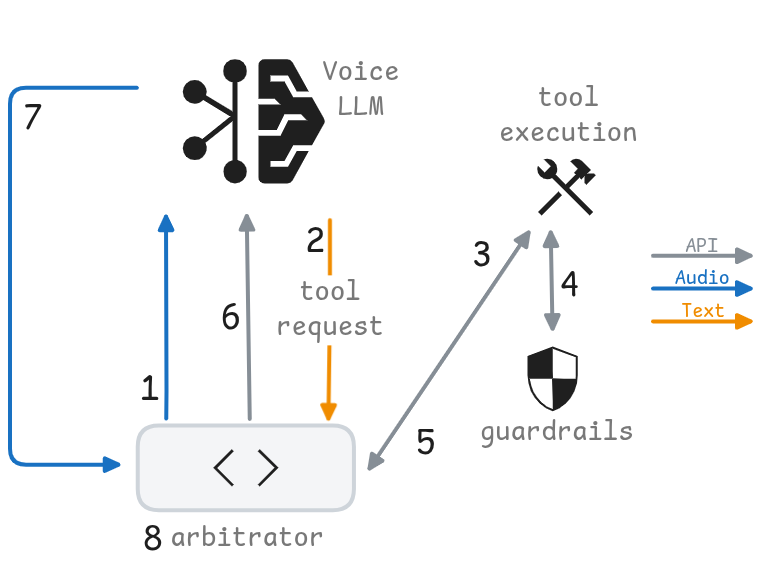}
  \vspace{-0.4cm}
  \caption{Tool-Based Guardrails}
  \label{figure:tool_guardrails}
\end{minipage}
\vspace{-0.4cm}
\end{figure}

Based on the considerations above, we implemented two variants of S2S guardrails: Transcript-based and tool-based.
Transcript-based guardrails denote functionality operating on transcripts handed back by the LLM.
The transcripts may be constructed by the main LLM itself (as is the case for Gemini Flash \cite{gemini-flash} and Nova Sonic \cite{nova-sonic-transcripts}) or by an auxiliary model (e.g. whisper-1 or gpt-4o-transcribe in case of GPT Realtime \cite{gpt-realtime-transcripts}).
Figure \ref{figure:transcript_guardrails} shows their data flow:
(1) An arbitrator (e.g. a car's software backend) sends audio to the LLM.
(2) In parallel, the LLM (a) creates a transcript that is passed to the input guardrails, and (b) creates an audio answer that is sent to the arbitrator and buffered there.
(3) The input guardrails notify the arbitrator about their result based on the transcript.
(4) The arbitrator sends the buffered audio to the output device if the guardrails were passed. Otherwise, the arbitrator instructs the model to react with previously defined refusal behavior.

Tool-based guardrails denote separate functionality whose execution is requested by the LLM.
Figure \ref{figure:tool_guardrails} shows the data flow:
(1) The arbitrator sends audio to the LLM.
(2) The LLM sends a tool request for the guardrails tool containing necessary parameters (like the transcript).
(3) The arbitrator passes on the request and parameters to the guardrails tool.
(4) The guardrails tool judges the input and passes back the result to tool execution.
(5) Tool execution notifies the arbitrator about the result.
(6) The arbitrator passes on the full tool result to the LLM.
(7) The LLM decides between blocking and passing behavior based on the guardrails tool result, generates fitting audio, and sends the audio to the arbitrator.
(8) The arbitrator sends the audio to the output device.
Model providers may offer configurable guarantees about deterministic tool usage \cite{amazon-speech-tools}.
However, at the time of this writing no vendor offered comprehensive tool pipelining which may be necessary if guardrails require several tools, or if guardrails coexist with other mandatory tools.

Note that while transcript-based guardrails show fewer communications steps than tool-based guardrails between components in the architecture, the former's implementation requires a larger amount of custom code since it offloads fewer steps to the LLM.

\section{Evaluation Setup\label{section:evaluating_guardrails}}
Detection accuracy or similar measures are not part of our analysis.
We focus on the following non-functional requirements since they are a necessary precondition for industrial application:
(R1) minimal added latency, (R2) deterministic model response when refusing user input, and (R3) safety first, i.e. guardrails must reliably be evaluated before every model answer.
R2 and R3 are general requirements for 
guardrailed systems, while R1 specifically concerns the conversation flow benefits S2S systems deliver over chained systems:
Adding delay as low as 0.4 seconds into conversations negatively impacts people's impression of their counterpart \cite{Schoenenberg2014} and introduces confusion into the conversation \cite{brady1971}.
Therefore, for S2S systems to be able to play to their strengths, guardrails introduced into the system should not significantly delay answer times.

Evaluation is broken up along the dimensions model (Google's Gemini Flash, Amazon's Nova Sonic, OpenAI's GPT Realtime), guardrails (none, tool-based, transcript-based), and input (malicious, non-malicious).
Malicious input is defined as all text containing the specified trigger word ``whale'', a benign word that model vendors' own safety checks should not identify as problematic.
This rules out conflicts between self-implemented and vendor-implemented guardrails.
The three test dimensions' value sets yield $3 * 3 * 2 = 18$ distinct setups.
Each test run was repeated 300 times, resulting in a total number of $5400$ runs.
To prevent temporary black box effects or time-related influences (like temporarily slow internet connectivity) from potentially overshadowing the effects of interest, the tests were repeated only after a single run along all test dimensions was completed, i.e. after each unique combinations of model, guardrails, and input.

\section{Evaluation Results \label{section:results}}
As overall latency, we track
the time that passes between the end of the user input (i.e. the time the user stops speaking) and the first audible response of the system.
While the tool-based approach means no answer is generated until the mandatory guardrail result is returned, transcript-based guardrails need to interrupt running answer generation.
At the time of this writing, not all vendors offered programmatic interruptions of a running session.
To preserve comparability between vendors, interruption was performed by streaming a fixed audio message to the model asking for refusal.
Duration of this audio message was subtracted from pertaining latencies to emulate an immediate interruption.
\begin{figure}[htb]
  \centering
  \vspace{-0.2cm}
  \includegraphics[width=0.999\linewidth]{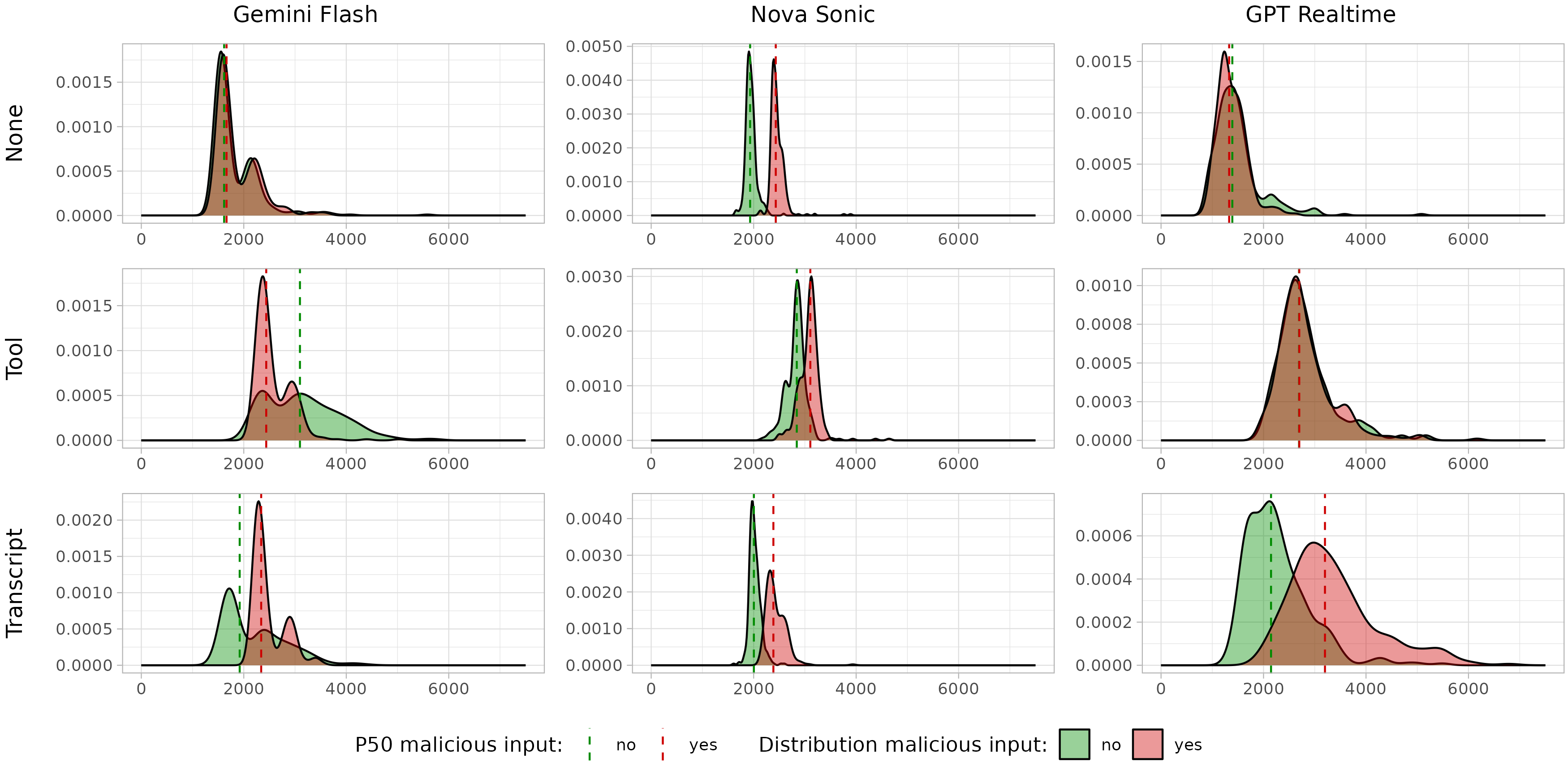}
  \vspace{-0.7cm}
  \caption{Guardrails and Models Comparison}
  \label{figure:latencies_overview}
  \vspace{-0.2cm}
\end{figure}

Figure \ref{figure:latencies_overview} shows the results.
The x-axis shows the delay between user speech end and start of the model's audio playback in milliseconds, the y-axis shows the density distribution of measurements, separated by whether the input was malicious or not.
Each combination of model and guardrail type is shown in a separate plot.
In addition to the density, the P50 quantiles for non-malicious and malicious inputs are shown by dashed lines.
Refer to table \ref{table:latencies_overview_table} for the numeric values of P50, P90, and P99 quantiles, in addition to mean and standard deviation (truncated to full milliseconds).

\begin{table}[htb]
\centering
% \vspace{-0.4cm}
  \caption{Latencies - Milliseconds non-malicious/malicious}
  \label{table:latencies_overview_table}
  \vspace{-0.2cm}
\begin{tabularx}{0.98\textwidth}{l|l|XXXXl}
  % \hline\hline
  model & guardrails\hspace{0.2cm} & P50 n/m& P90 n/m& P99 n/m& mean n/m& SD n/m\\ 
  \hline\hline
  Gemini\hspace{0.2cm} & none & 1616/1663 & 2278/2341 & 3598/3311 & 1828/1862 & 485/409 \\ 
  Gemini & tool & 3096/2438 & 4109/3044 & 5009/3566 & 3149/2574 & 723/356 \\ 
  Gemini & transcript & 1920/2339 & 3070/2960 & 4090/3421 & 2216/2477 & 602/313 \\ 
  \hline
  Nova & none & 1931/2431 & 2070/2591 & 2269/2876 & 1949/2457 & 127/155 \\ 
  Nova & tool & 2841/3106 & 3018/3265 & 3478/3447 & 2825/3081 & 220/213 \\ 
  Nova & transcript & 2004/2385 & 2153/2658 & 2333/2924 & 2022/2431 & 112/192 \\ 
  \hline
  GPT & none & 1391/1329 & 2112/1708 & 3035/2320 & 1488/1379 & 482/285 \\ 
  GPT & tool & 2693/2697 & 3545/3586 & 4766/5040 & 2819/2841 & 556/646 \\ 
  GPT & transcript & 2146/3199 & 3107/4591 & 4394/5811 & 2274/3377 & 627/866 \\ 
\end{tabularx}
\vspace{-0.5cm}
\end{table}

For the ``no guardrails'' case, the difference between malicious and non-malicious input is expectedly negligible.
Only Nova Sonic shows a slight difference, at a more narrow overall distribution than the other models.
This difference is likely explained by the mildly different user inputs.

Tool guardrails delay answers for all models by roughly 0.6 to 1.4 seconds on average, with marginally larger delays for Gemini Flash and GPT Realtime compared to Nova Sonic.
This impacts both malicious and non-malicious cases, as tool use always requires two model inferences (one to request the tool's result, another to use the tool's output) independent of the kind of input.
Gemini Flash exposes a broader distribution in the non-malicious case, and a few outliers with longer durations occurred for Gemini Flash and GPT Realtime.
Apart from these differences, and from the shift caused by added latency, the basic behavior of the models remain largely similar to the no guardrails case.

Transcript guardrails show 
slightly more varied distributions comparing malicious and non-malicious cases.
Overall, they outperform tool-based guardrails in every case except for GPT Realtime and malicious input.
The added latency in the non-malicious case is small enough for productive deployment for Gemini Flash and Nova Sonic.
The latter shows added latency around 0s for transcript guardrails in both cases.
Note that the non-malicious case is less impacted for Gemini Flash and Nova Sonic than for GPT Realtime.
This is likely due to the differing architecture of the corresponding inference systems, i.e. GPT Realtime's separate transcription model (see section \ref{sec:implementation_options}).

\begin{table}[ht]
\newcolumntype{C}{>{\centering\arraybackslash}p{1.54cm}}
\renewcommand{\arraystretch}{1.1}
\centering
\vspace{-0.2cm}
\caption{Requirements Compliance}
\label{table:compliance_matrix}
\vspace{-0.2cm}
\begin{tabularx}{0.95\textwidth}{|C|C|C|C|C|C|C|}
  \hline
  & \multicolumn{3}{c|}{\textbf{Transcript-based}} & \multicolumn{3}{c|}{\textbf{Tool-based}}\\
  \cline{2-7}
  & \textbf{Gemini} & \textbf{Nova} & \textbf{GPT} & \textbf{Gemini} & \textbf{Nova} & \textbf{GPT}\\
  \hline
  \textbf{R1 non.
  }  & \cellcolor[HTML]{A5D6A7} 0.3s & \cellcolor[HTML]{A5D6A7} 0s & \cellcolor[HTML]{FFE082} 0.7s & \cellcolor[HTML]{EF9A9A} 1.4s & \cellcolor[HTML]{FFE082} 0.9s & \cellcolor[HTML]{EF9A9A} 1.3s\\
  \hline
  \textbf{R1 mal.
  } & \cellcolor[HTML]{FFE082} 0.6s & \cellcolor[HTML]{A5D6A7} 0s & \cellcolor[HTML]{EF9A9A} 1.8s & \cellcolor[HTML]{FFE082} 0.7s & \cellcolor[HTML]{FFE082} 0.6s & \cellcolor[HTML]{EF9A9A} 1.3s\\
  \hline
  \textbf{R2
  }  & \cellcolor[HTML]{A5D6A7} ok & \cellcolor[HTML]{A5D6A7} ok & \cellcolor[HTML]{A5D6A7} ok & \cellcolor[HTML]{A5D6A7} ok & \cellcolor[HTML]{A5D6A7} ok & \cellcolor[HTML]{A5D6A7} ok\\
  \hline
  \textbf{R3
  }   & \cellcolor[HTML]{A5D6A7} ok & \cellcolor[HTML]{A5D6A7} ok & \cellcolor[HTML]{FFE082} limited & \cellcolor[HTML]{FFE082} limited & \cellcolor[HTML]{A5D6A7} ok & \cellcolor[HTML]{FFE082} limited\\
  \hline
\end{tabularx}
\vspace{-0.5cm}
\renewcommand{\arraystretch}{1}
\end{table}

The overall compliance of the analyzed model and guardrail options with the requirements from section \ref{section:evaluating_guardrails} is shown in table \ref{table:compliance_matrix}.
Checking for deterministic refusal response (R2) did not expose any adverse behavior in our tests.
Concerning safety, i.e. guardrails before answer (R3), transcript-based guardrails were well behaved with the (anecdotal) exception of some transcription errors for GPT Realtime, which could lead to guardrails erroneously passing.
While we observed no errors for tool-based guardrails concerning R3, only Nova Sonic strictly guarantees tool execution before query answer.
Concerning added latency (R1), all tool-based guardrails exceed the acceptable threshold with delays of 0.6-1.4 seconds for both benign and malicious input.
Transcript-based guardrails violate this requirement for malicious input for Gemini Flash, and for both input types for GPT Realtime.
Nova Sonic with transcript-based guardrails is the only combination that does not surpass the established 0.4s threshold, with almost no added latency, and is therefore fully compliant with R1.

\section{Future Work and Conclusion\label{section:conclusion}}
Speech2Speech assistants offer compelling features like natural conversation flow.
However, these come at a cost: the consolidation of speech processing into opaque, vendor-controlled models eliminates the intermediate text representations that conventional LLM guardrails depend upon.
This case study examines two implementation options for regaining this control.
The question it answers is ''can customizable guardrails be integrated into S2S systems such that S2S benefits are preserved''.
We find that only one of the three vendor offerings investigated satisfies the corresponding requirements at the time of this writing.
By consequence, productive use of S2S systems is currently dependent on external factors dictated by vendors, and thus is subject to change.
As a next step, studies on the effectiveness of customizable guardrails according to their specific use cases are a necessity.

For vehicle assistants, this means the requirements for production-grade S2S guardrails cannot yet be fully guaranteed.
Several developments could change this picture: vendor APIs exposing reliable and fast transcript streaming, mid-session refusal mechanisms, and novel approaches such as audio-native guardrails that bypass text intermediaries entirely. 
The automotive industry has long balanced innovation against safety.
S2S technology is no exception: its benefits are evident, but so are the gaps between current capabilities and production requirements.
Closing these gaps will require coordinated progress from model vendors, framework developers, and automotive integrators alike.

\bibliographystyle{splncs04}
\bibliography{s2s_guardrailing.bib}

\end{document}